\documentclass[preprint,5p,times,twocolumn,a4paper]{elsarticle}

\usepackage{csar}

\hypersetup{
  pdftitle={CSAR: Containerized System Architecture for Robotics},
  pdfkeywords={robotics architecture, containerization, multi-user infrastructures, LXC/LXD, ROS 2, DDS, edge--cloud continuum, resource isolation, reproducibility, GPU sharing},
  pdfsubject={Journal Paper}
}

\begin{document}

\begin{frontmatter}

\title{CSAR: Containerized System Architecture for Robotics\tnoteref{t1}}

\author[uma]{Gregorio Ambrosio-Cestero\corref{cor1}}
\cortext[cor1]{Corresponding author}
\ead{gambrosio@uma.es}
 
\author[uma]{Cipriano Galindo Andrades}
\ead{cgalindo@uma.es}
 
\author[uma]{Javier Gonzalez-Jimenez}
\ead{javiergonzalez@uma.es}
 
\author[uma]{Jose-Raul Ruiz-Sarmiento}
\ead{jotaraul@uma.es}
 
\address[uma]{Machine Perception and Intelligent Robotics group (MAPIR),
  Dept.\ of System Engineering and Automation,
  M\'alaga Institute for Mechatronics Engineering and Cyber-Physical Systems (IMECH.UMA),
  University of M\'alaga,
  Blvr.\ Louis Pasteur 35, 29071 M\'alaga, Spain.}

\begin{abstract}
Robotic applications increasingly rely on distributed computational infrastructures that combine embedded devices, edge servers, and cloud resources. This evolution, together with the collaborative nature of robotics projects, has made the development, integration, deployment, and long-term operation of robotic systems significantly more complex. In practice, multi-user robotics software teams face persistent challenges related to dependency isolation, compatibility, reproducibility, efficient sharing of specialized hardware, and deployment across heterogeneous environments. 
In this paper, we present CSAR (Containerized System Architecture for Robotics), a container-centric architectural framework designed specifically for robotics teams and the edge–cloud continuum. CSAR combines LXC/LXD-based system containerization, ROS~2/DDS-based communication, and a three-layer edge infrastructure to organize computation into hardware-affine, persistent execution environments that remain decoupled from the volatility of experimental workloads. Through its \textit{Infrastructure Core}, \textit{Platform and Multi-User Orchestration}, and \textit{Compute and Acceleration} layers, CSAR provides strong isolation, controlled resource sharing, and topology-aware networking for distributed robotic applications. 
To demonstrate its validity, we describe a real deployment of CSAR in an academic robotics laboratory and evaluate it through representative use cases involving edge-offloaded 3D SLAM and GPU-accelerated semantic mapping. The results indicate that CSAR simplifies software integration, improves the utilization of shared computational resources, and facilitates safe prototyping, as well as reproducible and collaborative experimentation in robotics teams.
The implementation described in this paper, including deployment templates, configuration files, and documentation, is publicly available at~\href{https://github.com/goyoambrosio/CSAR}{https://github.com/goyoambrosio/CSAR}.

\end{abstract}


\end{frontmatter}

\goyo{To comment on the text itself ...}

\cipri{To comment on the text itself ...}

\raul{To comment on the text itself ...}

\javier{To comment on the text itself ...}

\section{Introduction}

The rapid growth of robotic applications across diverse domains has significantly increased the complexity of the computational infrastructures that support their development, integration, deployment, and long-term operation.
Modern robotic solutions no longer consist of isolated programs running on individual machines, but of complex software stacks combining perception, mapping and localization, motion, task planning, semantic understanding, and decision-making, increasingly incorporating AI-driven pipelines based on large-scale models~\cite{vaswani2017_attention_all_need,radford2018_improving,devlin2019_bert,radford2019_language_models,openai2023_gpt_eb,moncada-ramirez2025_agentic_workflows}. These workloads often require high-performance hardware and distributed execution environments.
As a consequence, the management of robotic software-- typically carried out by multi-user teams--poses persistent challenges in compatibility, reproducibility, efficient sharing of computational resources, and deployment. Despite substantial progress in robotic middleware standardization, notably Robot Operating System (ROS~2)~\cite{macenski2022_robot_operating} and Data Distribution Service (DDS)~\cite{omg2015_omg_dds}, these frameworks define how nodes communicate, but not where computation should run, how it should bind to hardware, or how shared infrastructure should be organized and operated over time. Addressing these challenges therefore requires an architectural framework that relieves research teams from repeatedly assembling ad hoc system solutions.

Moreover, although robotic software development is inherently collaborative and long-lived, the computational infrastructure supporting it often remains fragmented and individual-centric. Robotics team members typically operates on a dedicated workstation, frequently equipped with costly GPU hardware and customized software configurations. Over time, this model generates recurrent structural tensions:

\begin{enumerate}
    \item \textbf{Infrastructure--deployment incoherence:} robotic applications are often developed in isolated personal environments but must later be integrated and executed on shared, heterogeneous infrastructure.
    \item \textbf{Hardware underutilization:} expensive computational resources remain partially idle and tightly coupled to specific users.
    \item \textbf{Reproducibility fragility:} software environments are implicitly bound to personal configurations, hindering long-term maintenance and integration.
    \item \textbf{Dependency and version friction:} heterogeneous requirements lead to fragile integrations and unresolved compatibility conflicts.

\end{enumerate}

These issues reveal a deeper problem: robotics laboratories still lack an explicit multi-user operating model that unifies software development, integration, deployment, and execution under consistent architectural principles. Although robotics has progressively adopted DevOps-inspired practices such as containerization, continuous integration, automated testing, and reproducible build environments~\cite{silva2023_devops_robotics, ronanki2021_robotic_software, mavrinac2023_devops_robotics}, these approaches primarily improve the engineering of software artifacts. They do not, by themselves, define how a shared computational infrastructure should be organized, allocated, and operated in environments where workloads depend on specific hardware, physical devices, and low-latency communication. This limitation is especially relevant in research laboratories, where multiple users must develop, integrate, and execute robotic applications over time on heterogeneous shared resources. Furthermore, most of these practices were originally conceived for cloud services and web-scale software systems rather than for embodied, time-sensitive systems interacting with the physical world. Consequently, robotics laboratories still lack a coherent operational model that connects these engineering practices with the management of the infrastructure on which robotic computation actually runs.

We address this gap through \textbf{CSAR (Containerized System Architecture for Robotics)}, an architectural framework designed specifically for robotics software management teams and the edge--cloud continuum. CSAR is not intended as a general-purpose cloud platform; instead, it provides an operating model for shared robotics infrastructure that restores coherence between robotic system design and the infrastructure in which those systems are developed and executed.

CSAR reinterprets multi-user computing under robotics constraints by treating containers as persistent execution environments rather than lightweight packaging artifacts, and by making placement, hardware affinity, and topology explicit architectural concerns.

\begin{figure}[t]
    \centering
    \includegraphics[width=0.9\columnwidth]{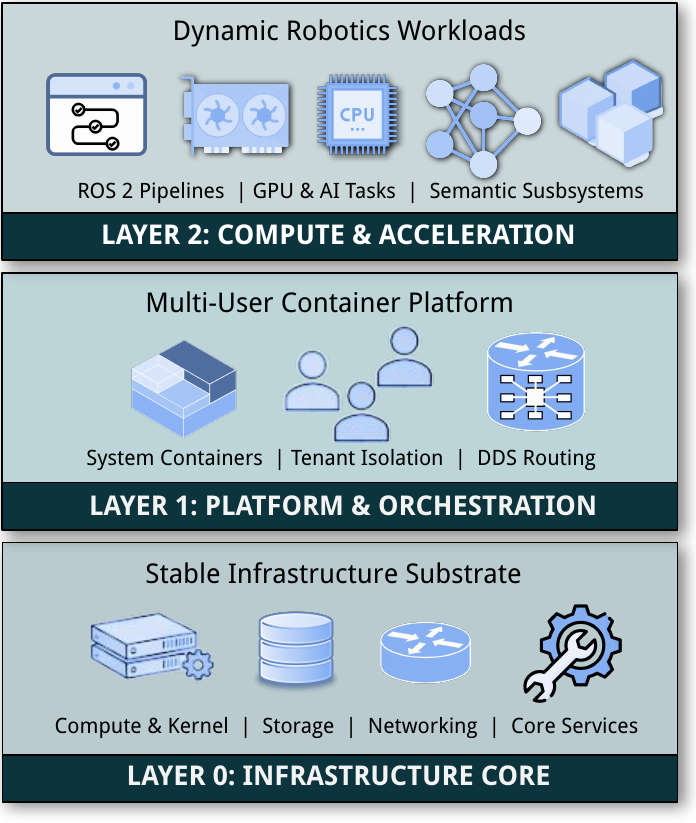}
    \caption{The CSAR architectural framework consists of layered multi-user infrastructure for robotics teams.}
    \label{fig:CSAR_FULL}
\end{figure}

The framework is structured into three functional layers
(see \FIG{\ref{fig:CSAR_FULL}}):

\begin{itemize}
    \item \textbf{Layer 0 --- Infrastructure Core:}
    a stable and deliberately conservative substrate responsible for kernel management, resilient storage, core networking, and long-lived services. This layer prioritizes predictability and longevity.

    \item \textbf{Layer 1 --- Platform \& Multi-User Orchestration:}
    the execution fabric built upon Linux system containers (LXC/LXD). This layer governs lifecycle management, network isolation, controlled interconnection, and cross-host portability of user environments.

    \item \textbf{Layer 2 --- Compute \& Acceleration:}
    the dynamic workload layer where ROS~2 nodes, robotic processing pipelines, and GPU-accelerated processes execute with near-native hardware access. This layer is intentionally disposable and isolated from infrastructure stability concerns.
\end{itemize}

By structuring the robotics infrastructure as a layered multi-user system rather than as a collection of personal machines, CSAR extends the logic of ROS~2 downward into infrastructure. It explicitly manages where computation lives, which hardware it binds to, and how failures are contained without compromising overall system stability. In this sense, CSAR provides an architectural framework that mediates between computation, resources, time, and physical locality.

The main contributions of this paper are the following aspects of the CSAR architectural framework:
\begin{itemize}
    \item \textbf{An explicit multi-user operating model for robotics infrastructure}, 
    which reconciles distributed robotic system design with shared infrastructure by organizing computation into hardware-affine LXC/LXD system containers instead of personal workstations or stateless microservices.
    \item \textbf{A layered architectural decomposition (Infrastructure Core / Platform / Compute)}
    that cleanly separates long-lived stability concerns from disposable research workloads, enabling failure containment, reproducibility, and controlled resource sharing without virtualization overhead.
    \item \textbf{A topology-aware container networking scheme aligned with ROS~2 and DDS semantics},
    where broadcast domains, discovery mechanisms, and cross-host communication are explicitly structured instead of implicitly inherited from generic cloud networking models.
\end{itemize}
The implementation described in this paper, including deployment templates, configuration files, and documentation, is publicly available at~\href{https://github.com/goyoambrosio/CSAR}{https://github.com/goyoambrosio/CSAR}.

In particular, CSAR has been successfully deployed in the Machine Perception and Intelligent Robotics (MAPIR) laboratory at the University of Málaga. To illustrate the framework in a realistic setting, we report practical aspects of that deployment and describe two representative use cases that highlight the key principles and capabilities of CSAR: the execution of an edge-offloaded 3D SLAM pipeline over 5G and the building of semantic maps in a distributed GPU-accelerated scenario.
Beyond the specific metrics reported in these scenarios, the deployment highlights several practical outcomes of the framework: safe prototyping of new robotic pipelines on shared infrastructure, controlled sharing of GPUs and sensing resources among multiple users, and improved reproducibility and long-term evolution of complex experiments without disrupting ongoing work.

The remainder of the paper is organized as follows. \SEC{\ref{sec:related_works}} reviews related work on virtualization, edge--cloud architectures, and multi-user robotic infrastructures. \SEC{\ref{sec:description}} presents the CSAR architectural framework. \SEC{\ref{sec:deployment}} describes its deployment in the MAPIR laboratory, and \SEC{\ref{sec:system_demonstration}} evaluates the framework through representative use cases.

\section{Related Works}
\label{sec:related_works}

The proposed CSAR architectural framework provides a coherent basis for the development, deployment and long term operation of modern robotic applications that require integrating heterogeneous software components, managing complex dependencies, and distributing computational workloads across edge and cloud infrastructures. This section reviews some of the ingredients of CSAR and compares the proposed approach with related works in the robotics literature.

\subsection{Virtualization and Containerization in Robotics}

To mitigate the dependency conflicts that frequently arise in complex software deployments \citep{lotfi2022_software}, developers have traditionally relied on hardware virtualization. However, traditional hypervisors introduce significant computational and latency overheads, making them less suitable for the real-time constraints of robotic systems \citep{soltesz2007_container}.
Consequently, lightweight virtualization via containers has become widespread in the robotics community. Docker \citep{merkel2014_docker} is widely adopted to package ROS environments, ensuring application consistency across heterogeneous machines \citep{white2017_ros}. Recently, containerized deployments have been increasingly adopted to build zero-setup, remote development environments for ROS~2 education and prototyping \citep{krumins2026_open_source}.
Despite its popularity, Docker is fundamentally an application container engine designed for ephemeral, single-process, and stateless microservices. Extensive performance evaluations of containerization in edge--cloud computing stacks highlight that while application containers are efficient for isolated tasks, they introduce structural complexities and overheads when simulating full operating system environments for industrial applications \citep{liu2021_performance_evaluation}. This is further corroborated by recent studies that quantify how mixing host and containerized ROS~2 deployments introduces measurable overheads in task periodicity, jitter, and end-to-end communication latency \citep{betz2025_end_end}. This stateless paradigm directly conflicts with the needs of a multi-user robotics infrastructure, which requires persistent, stateful workspaces.
In contrast, system containers based on LXC/LXD encapsulate a complete Linux distribution while sharing the host kernel \citep{graber2014_lxc}. As demonstrated by Plauth et al. \citep{plauth2017performance}, this lightweight approach to virtualization achieves near-native performance, avoiding the heavy abstraction layers of traditional hypervisors while providing the persistent, isolated workspaces required by robotics team members. While previous works explored system containers for managing robot software locally \citep{wang2019_rorg}, CSAR scales this concept into a comprehensive, multi-user infrastructure spanning the entire robotics infrastructure continuum.

\subsection{Distributed Robotics and Edge--Cloud Architectures}

As robotic workloads increasingly exceed the onboard processing capacity of the platforms that host them, researchers rely on distributed edge--cloud architectures \citep{tahir2025_edge_computing}. Frameworks such as FogROS~2 \citep{ichnowski2023_fogros2} have successfully enabled the offloading of computationally intensive ROS~2 nodes to cloud infrastructure.
Managing communication and scheduling across these distributed nodes, however, remains a critical bottleneck. In enterprise IT, Kubernetes \citep{rensin2015_kubernetes_scheduling} is the de facto standard for container orchestration. Recent efforts have attempted to apply Kubernetes to enhance the resilience of ROS~2-based multi-robot systems at the edge \citep{zhang2025_enhancing_resilience}. Yet, state-of-the-art orchestrators inherently struggle to support mixed-criticality systems and real-time robotic software \citep{lumpp2024_enabling_kubernetes}. As highlighted in recent surveys on real-time advancements in ROS~2, the custom internal scheduling mechanisms of ROS~2 executors and the multicast discovery of DDS \citep{omg2015_omg_dds} are heavily complicated by the network abstraction layers imposed by cloud-native platforms \citep{casini2025_survey_real}.
CSAR deliberately avoids the overhead of cloud-native orchestration platforms. Instead, it defines an explicit network topology that natively integrates DDS routers to bridge communication seamlessly between physical robots and LXC virtual environments. Furthermore, this bare-metal networking approach facilitates end-to-end latency optimizations that are crucial for containerized distributed autonomy \citep{betz2025_end_end}.

\subsection{Hardware Resource Sharing in Multi-User Contexts}

A major inefficiency in contemporary robotics is the underutilization of specialized hardware, particularly GPUs, which are essential for high-bandwidth sensory processing~\citep{baumann2024_cr3dt} and modern Artificial Intelligence models~\citep{moncada-ramirez2025_agentic_workflows}. In traditional setups, high-end GPUs are tied to individual workstations. While researchers have proposed transparent GPU sharing mechanisms in container clouds~\citep{wu2023_transparent_gpu}, implementing dynamic GPU multiplexing in cloud-native platforms often requires complex middleware. Recent studies emphasize that managing GPU executions securely and efficiently in real-time robotic contexts, such as autonomous driving, requires highly specialized scheduling to prevent priority collisions and execution delays~\citep{zhu2025_urgengo}.

CSAR addresses this complexity through Layer~2 of its architectural framework, leveraging LXC profiles to grant system containers direct, low-latency passthrough access to the host's physical GPUs. This allows multiple robotics team members to run hardware-accelerated tasks, such as CUDA-based perception pipelines, simultaneously and safely on the same underlying edge server with near bare-metal performance.

\section{CSAR Architecture}
\label{sec:description}

\begin{figure}[t]
  \centering

  \begin{subfigure}[b]{\columnwidth}
    \centering
    \renewcommand{\thesubfigure}{c} 
    \includegraphics[width=\columnwidth]{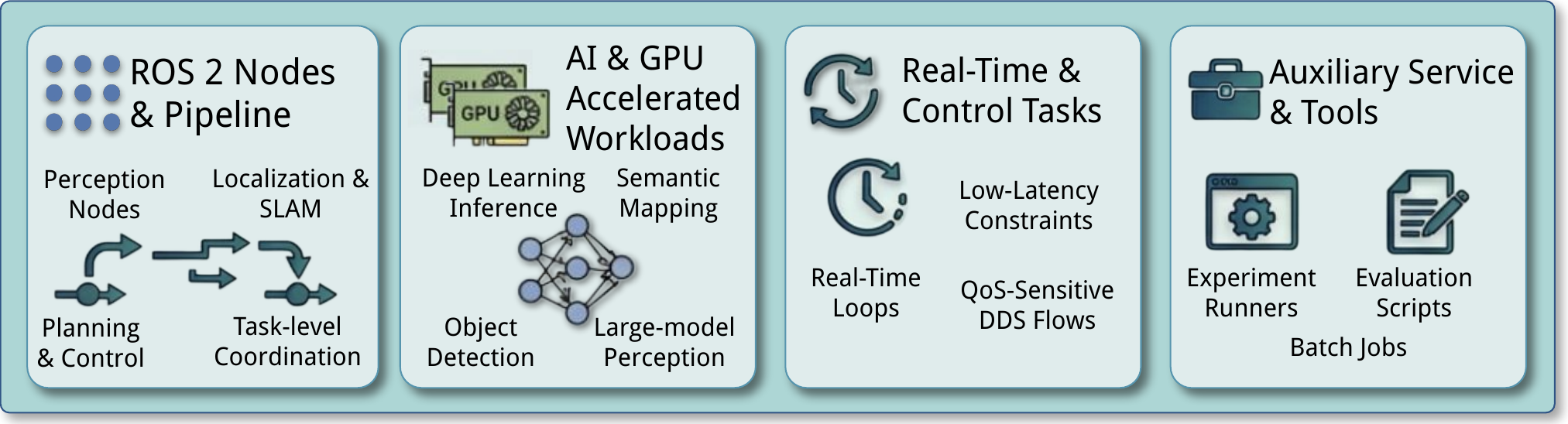}
    \caption*{\textbf{(c)} Layer 2: Compute \& Acceleration} 
    \label{fig:CSAR_LAYER_2}
  \end{subfigure}

  \vspace{0.4em}

  \begin{subfigure}[b]{\columnwidth}
    \centering
    \renewcommand{\thesubfigure}{b}
    \includegraphics[width=\columnwidth]{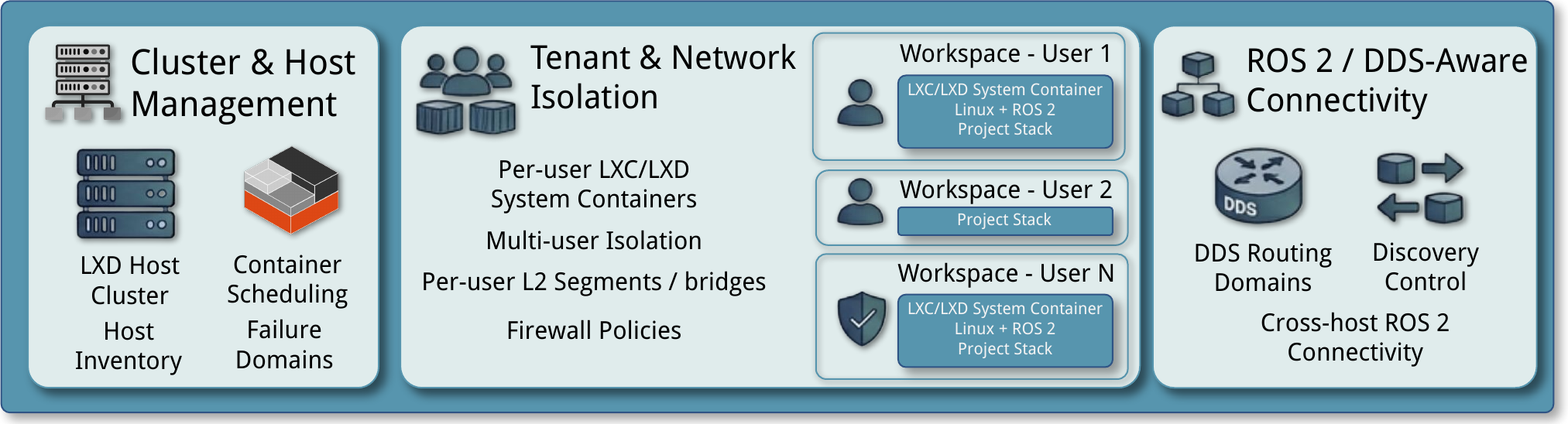}
    \caption*{\textbf{(b)} Layer 1: Platform \& Multi-User Orchestration}
    \label{fig:CSAR_LAYER_1}
  \end{subfigure}

  \vspace{0.4em}

  \begin{subfigure}[b]{\columnwidth}
    \centering
    \renewcommand{\thesubfigure}{a}
    \includegraphics[width=\columnwidth]{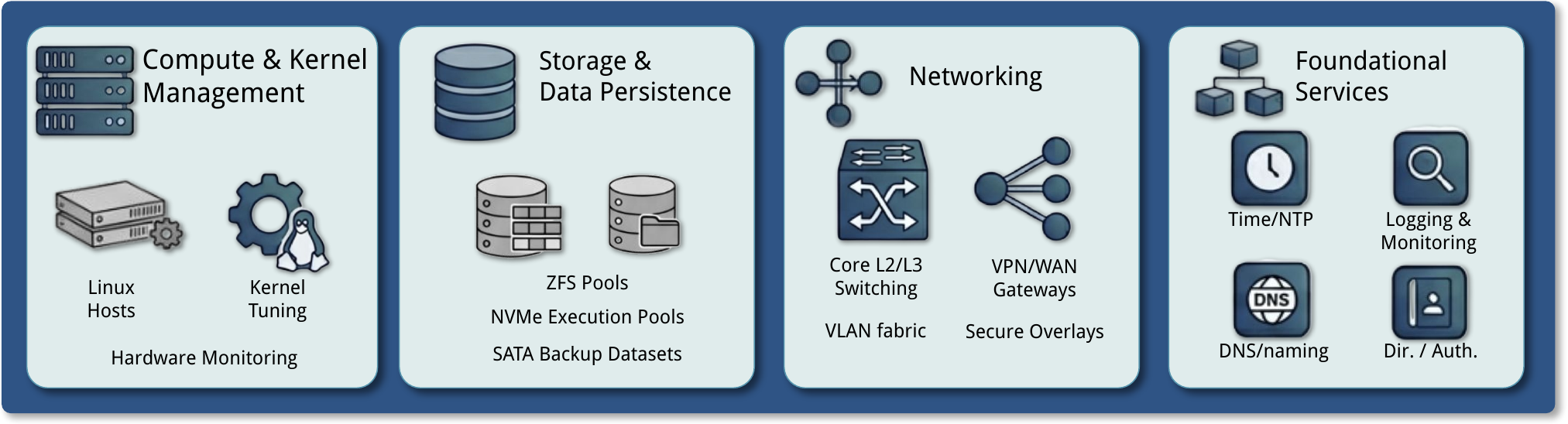}
    \caption*{\textbf{(a)} Layer 0: Infrastructure Core}
    \label{fig:CSAR_LAYER_0}
  \end{subfigure}

  \caption{Simplified representation of the main concepts within the three CSAR layers: (a) Layer 0, corresponding to the substrate for shared robotics infrastructure; (b) Layer 1, corresponding to LXC/LXD-based execution fabric and multi-tenant isolation; and (c) Layer 2, corresponding dynamic robotics workloads with near-native hardware access.}
  \label{fig:CSAR_LAYERS_PORTRAIT}
\end{figure}

This section presents the CSAR architectural framework, structured into three layers (see~\SEC{\ref{subsec:arch_principles}}). We describe the responsibilities of each layer and how they interact to provide isolation, reproducibility, and low-latency execution for distributed robotic workloads (see~\SEC{~\ref{subsec:layer0}--\ref{subsec:layer2}}). We then summarize the practical mechanisms that support stable operation, including monitoring and resource observability (see~\SEC{\ref{subsec:layer_interaction}}).

\subsection{Architectural Principles}
\label{subsec:arch_principles}

CSAR structures the computational infrastructure of a robotics environment as a layered system in which long-lived infrastructure services are clearly separated from dynamic robotic workloads. This separation bridges the gap between the computational demands of modern robotics and the physical constraints of shared laboratory infrastructures. The goal is to separate the stability of the physical hardware from the volatility of experimental software.

The design follows three guiding principles: first, infrastructure must remain stable over long periods of time; second, execution environments must be isolated but persistent; and third, robotic workloads must retain direct access to specialized hardware and low-latency communication paths. To support these requirements, CSAR decomposes the system into three functional layers: an infrastructure core (see~\SEC{\ref{subsec:layer0}}), a platform layer responsible for container lifecycle management (see~\SEC{\ref{subsec:layer1}}), and a dynamic workload layer where robotic applications are developed and executed (see~\SEC{\ref{subsec:layer2}}).

\subsection{Layer 0: Infrastructure Core}
\label{subsec:layer0}

The foundational layer, referred to as the Infrastructure Core, is deliberately designed to remain highly stable and conservative. Typical responsibilities of the infrastructure core include the management of storage pools, persistent datasets, and snapshot mechanisms, as well as the configuration of core networking services such as DNS, time synchronization, and secure connectivity between nodes. These elements, summarized in 
\FIG{\ref{fig:CSAR_LAYERS_PORTRAIT}\subref{fig:CSAR_LAYER_0}},
provide the fundamental operating conditions under which the rest of the system executes.
Another key responsibility of this layer is hardware exposure. Physical resources such as CPUs, GPUs, and high-speed network interfaces are made available to the upper layers through the host operating system. 
The infrastructure layer therefore determines the physical capabilities of the system but does not dictate how those resources are functionally applied. At this foundational level, the host operating system is strictly responsible for resource administration and allocation. The specific task execution, software configurations, and functional use of these resources are instead dictated by the dynamic robotic workloads deployed in Layer~2. Within Layer~0, the host maintains the following essential services:

\begin{itemize}
    \item \textbf{Core Networking and Routing:} Handled by a dedicated gateway router that bridges physical laboratory networks (Ethernet and Wi-Fi) with the virtual container domains, providing the L3 routing and policy boundary between physical networks and CSAR tenant subnets.
    \item \textbf{Secure Wide Area Networks (WAN):} Layer~0 hosts a private controller for a Secure Network Overlay (based on ZeroTier). This allows external physical robots or remote researchers to connect to the laboratory infrastructure via encrypted tunnels, providing authenticated remote membership for devices located outside the laboratory perimeter.
    \item \textbf{Layered Communication Support:} At Layer~0, the physical router and the overlay network provide secure IP reachability and enforce network-level access policies. Layer~1 then builds on this routable reachability to implement DDS discovery and routing services (Discovery Server and DDS Router) across segmented domains when ROS~2 communication must span users, hosts, or sites.
    \item \textbf{Resilient Storage (ZFS):} Persistent storage is managed through ZFS pools. The architecture  segregates high-speed NVMe solid-state drives, dedicated to the active execution of containers, from high-capacity SATA hard drives, which are reserved for system backups, container snapshots, and large shared datasets (such as sensory logs or ROS bags).

\end{itemize}

Because of its foundational nature, if Layer 0 experiences a critical failure, the entire multi-user ecosystem becomes unavailable; therefore, human intervention at this level is strictly limited to infrastructure administration.

\subsection{Layer 1: Platform and Multi-User Orchestration}
\label{subsec:layer1}

Layer~1 operates directly above the infrastructure core and constitutes the execution fabric of the architectural framework. It provides the mechanisms required to create, manage, and interconnect the containerized environments in which users deploy their robotic software stacks. These functions, summarized in 
\FIG{\ref{fig:CSAR_LAYERS_PORTRAIT}\subref{fig:CSAR_LAYER_1}},
are implemented through Linux system containers managed by LXC/LXD.

Unlike application-level containers (e.g., Docker), which typically encapsulate a single stateless process, system containers provide complete operating system environments while sharing the host kernel. This approach enables near-native performance and direct access to hardware resources while avoiding the computational overhead typically associated with traditional hardware hypervisors.

An important feature of this platform layer is its explicit support for multi-user operation. The orchestration system manages the lifecycle of containers, including their creation, replication, migration, and removal, while supporting cross-host portability. Consequently, each robotics team member operates within independent persistent containers that can be freely configured with different Linux distributions, library versions, and software dependencies. This strict separation eliminates many of the compatibility conflicts that commonly arise in shared robotics development environments.

In addition to the isolation of the environment, the platform layer is responsible for providing network isolation and controlled connectivity. ROS~2 communication is implemented through DDS, and interoperable DDS deployments typically use DDSI-RTPS for discovery and data exchange. Default DDS discovery (e.g., RTPS participant and endpoint discovery) often depends on multicast reachability within an L2 broadcast domain. In multi-user infrastructures, where containers are segmented across subnets and multicast may be filtered by routers, Wi-Fi isolation, or VPN boundaries, this can prevent nodes from discovering each other and, therefore, inhibit topic and service connectivity.

CSAR addresses this by separating (i) tenant isolation from (ii) cross-domain communication. Layer~1 provisions an isolated L2 domain per user to prevent cross-user interference, while cross-domain or cross-site ROS~2 communication is enabled through explicit DDS discovery and routing services (Discovery Server and DDS Router), which bridge only the selected ROS~2 interfaces required for controlled collaboration.

Finally, to ensure that distributed robotic components can interact transparently across container boundaries and physical locations, Layer~1 implements a Secure Network Overlay (SNO). Using a private controller hosted within the infrastructure, external physical robots and remote computers can join the virtual network through encrypted tunnels. When controlled collaboration is required, such as sharing sensor data between an outdoor robot and an edge container, the architectural framework uses DDS Routers and Discovery Servers to selectively bridge ROS~2 topics across the isolated subnets.

Although different DDS vendors provide different implementations, they interoperate at the network level through the standardized DDSI-RTPS protocol. CSAR therefore considers middleware selection a deployment parameter while keeping the communication architecture unchanged.

\subsection{Layer 2: Compute and Acceleration}
\label{subsec:layer2}

As illustrated in 
\FIG{\ref{fig:CSAR_LAYERS_PORTRAIT}\subref{fig:CSAR_LAYER_2}},
Layer~2 represents the user-facing dynamic workload environment. In contrast to the conservative nature of the infrastructure core, this layer is intentionally disposable and highly volatile. Here, robotics team members instantiate their customized development environments to execute complex robotic pipelines, such as real-time sensor fusion, semantic mapping, or Large Language Model (LLM) inference.

In this layer, containers act as robotic semantic subsystems rather than simple application packages. As modern robotic systems are inherently distributed, multiple containers frequently cooperate to implement a single robotic capability. For example, computationally heavy perception modules can run on GPU-equipped edge servers, while latency-critical control nodes execute closer to the physical robot platform. The layered architecture allows these distributed deployments to be organized transparently, maintaining consistent communication through the ROS~2 and DDS integration provided by Layer~1.

Containers in this layer are typically instantiated from preconfigured base images from Layer 1, which provide ready-to-use environments for robotics development. From the team members perspective, each instantiated container behaves as an independent, fully featured Linux machine in which they hold administrative (\textit{root}) privileges. This autonomy eliminates the dependency conflicts found in shared machines, allowing users to freely modify library versions or middleware configurations without affecting the stability of the underlying system.

Layer~2 also bridges the gap between isolated user environments and accelerated execution by enabling workloads to leverage the specialized hardware resources provided by the infrastructure core with minimal virtualization overhead. 
For example, using GPU-enabled base images with CUDA support, containers can access the available physical NVIDIA GPUs with near-native performance. This allows deep learning algorithms to exploit hardware acceleration within isolated user environments. In addition, because these environments are fully encapsulated,  robotics team members can capture snapshots of their containers before applying risky software changes, thus supporting reproducibility and safe experimentation.

\subsection{Cross-Layer Interaction and Failure Containment}
\label{subsec:layer_interaction}

While each architectural layer is strictly delimited by design, CSAR operates as a cohesive system in which the layers continuously negotiate the distribution of physical and virtual resources. Rather than behaving as a traditional stack of independent IT services, this hierarchy functions collectively to mediate between computation, hardware resources, and physical reality.

The interaction follows a clear bottom-up flow of capabilities. The infrastructure core (Layer 0) exposes the raw physical and networking resources, establishing the foundation of what is possible with hardware. The platform tier (Layer 1) acts as the execution fabric, translating these raw capabilities into isolated, multi-user domains. Finally, the compute tier (Layer 2) consumes these domains, providing the actual virtual environments where robotic workloads interact via middleware communication mechanisms.
In summary, the infrastructure core exposes capabilities, the platform layer organizes them into isolated domains, and the compute layer consumes them to execute robotic workloads.

Crucially, this structural separation addresses one of the core challenges in shared robotics infrastructure: failure containment. Traditional infrastructures often treat failure as an exception, whereas CSAR treats it as a structural reality that must be contained. By isolating the stable host environment from experimental workloads, errors remain strictly confined within their specific boundaries. For example, if a researcher introduces dependency conflicts or causes a fatal crash within a complex perception pipeline, the failure remains entirely isolated within the disposable compute container. The core infrastructure services continue operating unaffected, ensuring that other researchers sharing the same physical GPUs or server can continue working without interruption.

Conversely, infrastructure administrators can perform network reconfigurations, manage ZFS storage pools, or upgrade physical hardware at the core layer without altering the persistent software environments customized by the researchers. Ultimately, this layered interaction guarantees both operational stability and experimental flexibility. It addresses the long-standing tension between maintaining a reliable, long-lived robotics infrastructure and supporting the aggressive, error-prone iteration cycles required to advance modern robotics.

This section has described the conceptual architecture of CSAR. The following section presents its practical deployment within a real robotics development team, detailing the physical infrastructure and hardware resources that support the CSAR framework. 

\section{CSAR Deployment in the MAPIR Laboratory}
\label{sec:deployment}

The CSAR architectural framework is currently deployed within the Machine Perception and Intelligent Robotics (MAPIR) laboratory at the University of Málaga. Before its adoption, the MAPIR laboratory relied on a more traditional model based on individual workstations and ad hoc server configurations. Over time, this approach led to recurring problems in dependency management, GPU underutilization, and limited reproducibility across experiments.  

The infrastructure is organized as a small edge-computing cluster designed to support multiple researchers developing and deploying robotic applications simultaneously. The system enables these users to share high-performance computational resources while maintaining isolated execution environments. The following subsections describe the physical infrastructure (see~\SEC{\ref{subsec:physical_infrastructure}}), the multi-user operating model (see~\SEC{\ref{subsec:multi-user_operation}}), and the monitoring mechanisms (see~\SEC{\ref{subsec:monitoring}}) that support the stable operation of the deployment.

\subsection{Physical Infrastructure}
\label{subsec:physical_infrastructure}

The system is primarily hosted on two physical servers, referred to as \textit{edge} and \textit{uedge}. These machines constitute the computational backbone of the CSAR infrastructure and host the containerized execution environments described in Section~\ref{sec:description}. Together, they realize the hardware exposure and execution capacity discussed in Layer~0, providing different levels of computational power, shared storage, networking services, and access to hardware accelerators. In practice, this heterogeneous edge-computing tier combines a general-purpose entry node with a higher-end server for computationally intensive workloads, allowing the deployment to support both routine development tasks and GPU-accelerated robotic processing.

The \textit{edge} node serves as the primary entry point for general-purpose robotic workloads and container management. It is built on an Intel Core i7-11700K processor (8 cores at 3.60\,GHz), 128\,GB of DDR4 RAM, and two NVIDIA GPUs: an RTX~3060~Ti with 8\,GB of memory and a TITAN~X with 12\,GB of memory. In terms of storage, \textit{edge} incorporates a 1\,TB NVMe drive for the operating system, a 4\,TB NVMe drive for active container execution, and a 10\,TB SATA drive for persistent storage. The node is connected through a 2.5\,Gb Ethernet interface and is typically used for moderate workloads, although its GPUs can also be allocated to containers when additional acceleration is required.

The \textit{uedge} node is dedicated to computationally intensive workloads such as deep learning inference, perception pipelines, and dense 3D reconstruction. This server is equipped with an AMD Ryzen Threadripper PRO 7975WX processor (32 cores at 4.0\,GHz, with 162\,MB of cache), 512\,GB of DDR5 RAM, and three NVIDIA RTX~6000 Ada GPUs with 48\,GB of GDDR6 memory each. It also provides three 10\,Gb Ethernet interfaces, two connected to the laboratory network and one reserved for out-of-band remote management. In terms of storage, \textit{uedge} incorporates a 2\,TB NVMe drive for the operating system and CSAR images, a 4\,TB NVMe scratch drive used for active container execution, and an 18\,TB SATA drive used for repositories, backups, datasets (\ie Robot@Home2~\cite{goyo2023softwarex}), and other shared resources. Through LXC/LXD configuration profiles, GPU devices can be selectively exposed to containers, enabling robotic applications to access hardware acceleration while maintaining strict user isolation.

Connectivity between the servers and the robotic platforms deployed in the laboratory is provided through a hybrid networking infrastructure centered on a MikroTik hAP~ax\textsuperscript{3} router. This device combines four Gigabit Ethernet ports, one 2.5\,Gb Ethernet port, and Wi-Fi~6 connectivity, allowing the integration of wired laboratory resources with wireless robotic platforms. In addition to interconnecting the physical infrastructure, the router supports the low-latency communication required by distributed robotic systems operating both within and beyond the laboratory environment.

\subsection{Multi-User Operation}
\label{subsec:multi-user_operation}

The CSAR infrastructure is designed to support simultaneous use by multiple researchers. Each user accesses the system through a personal account on the host servers and can instantiate and manage their own containerized environments.

Containers are typically instantiated from preconfigured base images that provide ready-to-use development environments for robotics applications, including ROS~2 middleware and GPU-enabled software stacks (recall~\SEC{\ref{subsec:layer2}}). From the user's perspective, each container behaves as an independent Linux system with administrative privileges, allowing researchers to customize software dependencies and experimental configurations without affecting other users.

Each user is assigned a dedicated IPv4 subnet within the internal CSAR virtual network. In the current deployment, subnets follow the addressing scheme \texttt{10.<server>.<id>.0/24}, where \texttt{<server>} identifies the physical host (e.g., \texttt{1} for \textit{edge} and \texttt{2} for \textit{uedge}) and \texttt{<id>} corresponds to the numerical identifier of the user in the host system. Containers instantiated by a given user receive addresses within this subnet and are automatically registered in the internal DNS service using a Fully Qualified Domain Name (FQDN) structure of the form:
\begin{center}
\texttt{<container>.<user>.<server>.mapir}
\end{center}
For example, containers belonging to the user \texttt{foo} on the \textit{edge} server may receive addresses in the subnet \texttt{10.1.15.0/24}, while containers belonging to the same user on the \textit{uedge} server may use \texttt{10.2.15.0/24}.

This addressing scheme allows containers belonging to the same user to communicate transparently while preserving isolation between independent experiments. At the same time, controlled communication between containers belonging to different users can be enabled when collaborative experiments require it.

At the time of writing, the deployment supports 250 registered users. Concurrency varies with the time of day and the demand for GPU-intensive workloads: during regular periods, we observe roughly 15--25 concurrently running containers, while peak usage reaches approximately 35 when multiple GPU-accelerated experiments are executed simultaneously. These numbers show that the infrastructure is not merely a conceptual framework, but an actively used multi-user environment that supports daily operation in the laboratory.

\subsection{Monitoring and Resource Observability}
\label{subsec:monitoring}

To support stable operation of the shared infrastructure, CSAR incorporates a monitoring framework that provides real-time visibility into system resources and container workloads. This capability is particularly important in robotic systems, where variations in CPU load, GPU memory usage, or network latency can directly affect the performance of perception pipelines and control processes.

\begin{figure}[t]
    \centering
    \includegraphics[width=1.0\columnwidth]{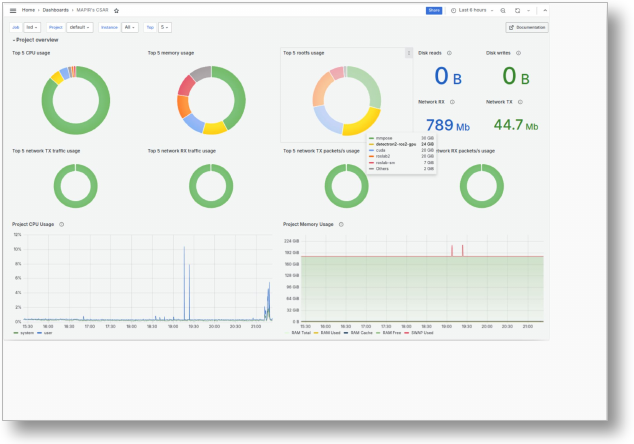}
    \caption{Overview of CSAR’s monitoring dashboards implemented with Grafana, showing container-level CPU, memory, GPU, and network utilization across the edge servers during robotic operation.}
    \label{fig:CSAR_GRAFANA}
\end{figure}

The monitoring stack is based on \textit{Prometheus} \cite{brazil2018_prometheus}, which collects time-series metrics from both the host systems and the LXD container environments. These metrics are visualized through \textit{Grafana} \cite{kiresova2023_grafana_visualization} dashboards (see \FIG{\ref{fig:CSAR_GRAFANA}}), allowing researchers and administrators to observe computational load, GPU utilization, thermal behavior, and energy consumption across the \textit{edge} and \textit{uedge} nodes.

In addition, CSAR provides a lightweight command-line interface that simplifies common operational tasks such as container lifecycle management, ROS~2 environment configuration, and hardware status inspection. This tooling reduces the operational complexity of the underlying container and networking infrastructure, allowing robotics team members to focus on the development and evaluation of robotic algorithms while maintaining awareness of the system’s operational state.

The CSAR infrastructure is currently used by several researchers in the MAPIR laboratory to develop and evaluate distributed robotic systems, ranging from perception pipelines to multi-robot experiments, as shown in the following section.

\begin{figure}[t]
    \centering
    \includegraphics[width=0.7\columnwidth]{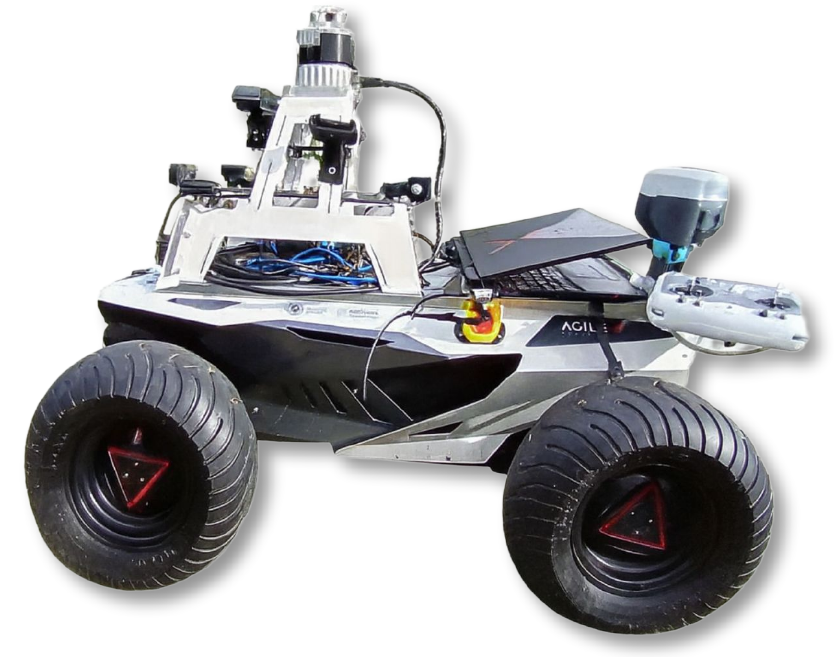}
    \caption{The Hunter Robot equipped with an Ouster 3d Laser, RTK GPS Receiver, and 3 Full HD cameras.}
    \label{fig:CSAR_HUNTER}
\end{figure}

\begin{figure}[t]
    \centering
    \includegraphics[width=1.0\columnwidth]{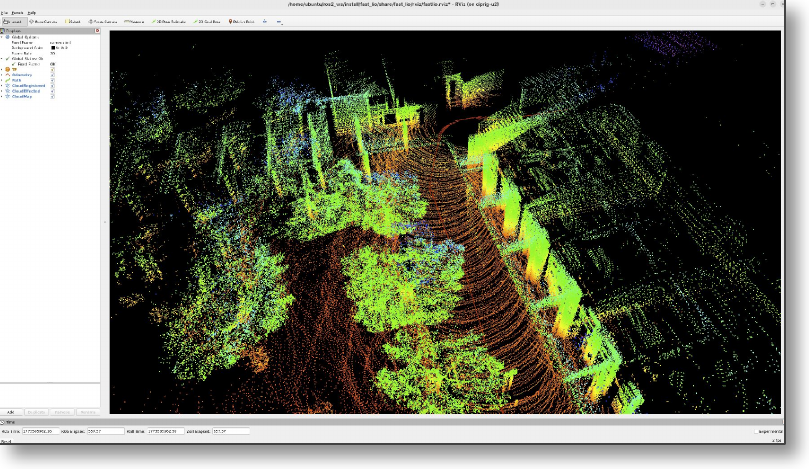}
    \caption{RViz visualization of the experimental output produced in the CSAR use case 1 setup. The point cloud map and estimated trajectory are displayed remotely from the visualization container running on the Uedge server, showing the 3D reconstruction generated from the replayed sensor data of the Hunter platform dataset.}
    \label{fig:CSAR_RVIZ}
\end{figure}

\begin{figure*}[t]
    \centering
    \includegraphics[width=0.9\linewidth]{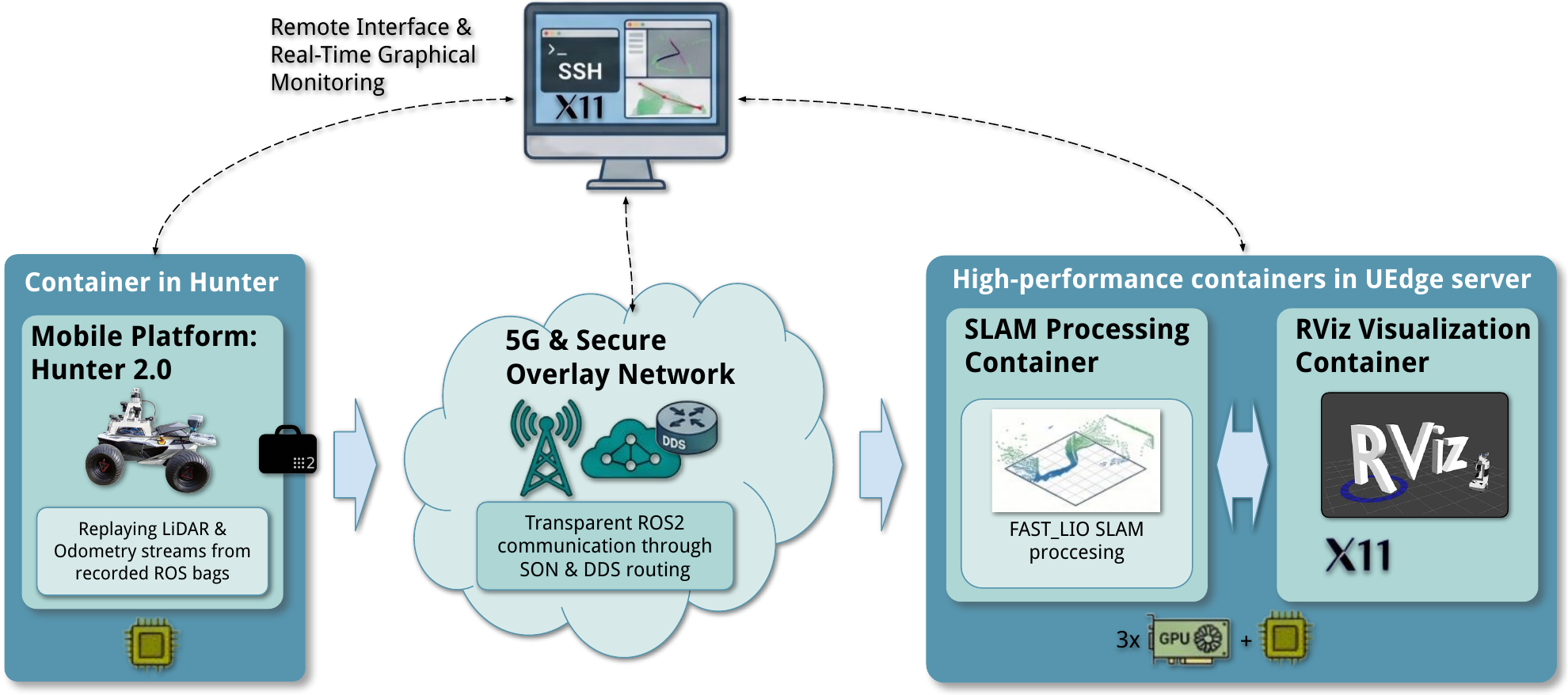}
    \caption{Use case 1 of the CSAR architecture. A ROS bag dataset emulates the sensor streams of the Hunter robotic platform and is replayed on a laptop connected through a 5G modem and secure overlay network. The data are transmitted to the Uedge edge server, where a container performs processing and SLAM while a second container visualizes the results using RViz. The researcher controls the experiment remotely via SSH and receives the graphical output through X11
forwarding.}
    \label{fig:CSAR_USE_CASE_1}
\end{figure*}

\section{System Demonstration}
\label{sec:system_demonstration}

In this section, we present two representative use cases to illustrate the operation of CSAR in a real deployment. The objective is not only to show that the proposed architectural framework can host distributed robotic workloads, but also to demonstrate how it supports the separation of data acquisition, computation, and visualization across the edge--cloud continuum while preserving ROS~2/DDS interoperability. Overall, these use cases also highlight how CSAR facilitates not only the final deployment but also the preliminary development and validation of robotic algorithms, allowing researchers to test configurations and resolve integration issues within the same infrastructure. The first use case focuses on edge-offloaded 3D SLAM over 5G (see~\SEC{\ref{subsec:use_case_1}}), whereas the second addresses distributed GPU-accelerated semantic mapping with heterogeneous software and hardware requirements (see~\SEC{\ref{subsec:use_case_2}}). Finally, we also provide some additional scenarios where the capabilities of CSAR are being leveraged (see~\SEC{\ref{subsec:additional_uses}}). 

\subsection{Use Case 1: Edge-offloaded 3D SLAM over 5G}
\label{subsec:use_case_1}

A common situation in mobile robotics is that the robot must acquire and transmit large volumes of sensory data while executing computationally demanding algorithms such as 3D SLAM. In many practical settings, equipping the robot with a high-end onboard computer is either undesirable or unnecessarily costly. This motivates the use of nearby edge resources, provided that communication and system integration remain sufficiently robust.

To demonstrate this scenario, we consider a distributed deployment in which the robot-side workload is limited to sensor playback and communication, while the SLAM pipeline is executed remotely on the CSAR infrastructure.

\subsubsection{Experimental setup}

The experiment emulates a mobile robot performing large-scale outdoor mapping. Rather than using the physical platform online, we rely on a pre-recorded dataset in order to ensure repeatability and facilitate reproduction of the experiment. (see \FIG{\ref{fig:CSAR_USE_CASE_1}})

The dataset was acquired with an AgileX Robotics Hunter 2.0 mobile platform equipped with an Ouster OS1-32 LiDAR, an RTK-GPS receiver, and two Full HD cameras (see \FIG{\ref{fig:CSAR_HUNTER}}). It corresponds to an approximately 1~km trajectory around the Computer Science School at the University of M\'alaga. The main ROS~2 bag corresponds to approximately 2.2\,GiB of recorded sensory data distributed as a 1.4\,GB compressed archive, complemented by a lightweight metadata bag required to decode and reconstruct LiDAR point clouds when needed. The recording covers approximately 15\,minutes of continuous robot motion. 

To support reproducibility, the dataset, including the corresponding rosbag files and the metadata required to decode the 3D point clouds, has been published in Zenodo~\cite{anaya2025_mobile_robot}.

\subsubsection{Distributed execution}

To evaluate the behavior of the CSAR architectural framework in a realistic wide-area setting, the dataset was replayed from a standard laptop acting as the remote robotic node. This laptop was not connected to the laboratory through the local network, but through a commercial 5G connection. In this way, the experiment reproduces a situation in which the robot operates outside the local infrastructure while still relying on remote computational support.

The workload was distributed into three containers managed within CSAR:

\begin{itemize}
    \item \textbf{Container 1 (sensor playback):} deployed on the remote laptop. This container replayed the recorded LiDAR and odometry streams through ROS~2.
    \item \textbf{Container 2 (SLAM processing):} deployed on the \textit{uedge} server. It subscribed to the incoming sensor streams and executed the FAST\_LIO pipeline~\citep{xu2021_fast_lio} for localization and map generation.
    \item \textbf{Container 3 (visualization):} also deployed on \textit{uedge} server. This container executed RViz for online monitoring of the reconstructed map. See \FIG{\ref{fig:CSAR_RVIZ}}
\end{itemize}

This partition reflects a typical CSAR deployment logic: lightweight acquisition close to the robot, computationally intensive processing on edge resources, and auxiliary tools such as visualization isolated in independent execution environments.
During the formulation of the experiment, this strict isolation proved highly beneficial for algorithmic design. For instance, researchers could repeatedly restart the SLAM processes in Container~2 to tune parameters without disrupting the ROS~2 bag playback in Container~1 or the RViz visualization in Container~3. This decoupling provided a stable framework for the continuous refinement of the algorithms.

\subsubsection{Communication support}

A relevant difficulty in this experiment was the transport of ROS~2/DDS traffic across the 5G connection. In particular, DDS discovery and data exchange are usually tied to local network assumptions that do not hold naturally across heterogeneous and geographically separated links.

To make this communication possible, we relied on the secure overlay networking mechanisms available in the lower layers of CSAR. More specifically, the remote node and the laboratory infrastructure were connected through a secure virtual overlay, and DDS traffic was bridged through a Discovery Server and DDS Router configuration. This allowed us to preserve transparent ROS~2 communication between containers located in different network segments, including the remote laptop and the isolated execution domains hosted in \textit{uedge}.

From the application perspective, this networking support allowed the distributed ROS~2 graph to operate as a single logical system despite the physical separation between the acquisition node and the processing server.

\subsubsection{Results and discussion}
The experiment shows that the 3D SLAM pipeline can be executed remotely on the edge infrastructure while the data source remained outside the laboratory and connected only through 5G. In this configuration, the remote node remained lightweight, whereas the most computationally demanding task was transferred to a more suitable execution environment. The dataset used in this experiment is non-trivial both in spatial scale and data volume. 
This combination is representative of bandwidth-demanding field deployments and motivates the need for explicit wide-area networking support.

The main result of this use case is therefore architectural rather than algorithmic. It shows that CSAR can sustain a distributed robotic workload in which (i) sensing, processing, and visualization are deployed in separate containers, (ii) communication spans heterogeneous networks, and (iii) ROS~2-based applications remain operational without requiring ad hoc modifications to the software stack.

At the same time, this experiment also highlights the practical importance of explicit network support in distributed robotics. Offloading computation is not only a matter of placing containers on more powerful machines; it also requires a communication substrate capable of coping with non-local links, segmented broadcast domains, and middleware discovery constraints. In our case, CSAR provides this support as part of the architectural framework (secure overlay membership plus Discovery Server/DDS Router), which is precisely what makes this type of deployment feasible in practice.

From the perspective of the CSAR architectural framework, this use case also illustrates the coordinated role of the three layers. Layer~0 provides the secure networking substrate that enables connectivity between the remote node and the laboratory infrastructure. Layer~1 supplies the isolated container environments and the DDS discovery and routing mechanisms required for controlled ROS~2 interoperability across segmented domains. Layer~2 hosts the distributed workloads themselves, separating sensor playback, SLAM processing, and visualization into independent execution environments. 

This organization not only supports the experiment technically, but also shows how CSAR facilitates the practical deployment of distributed robotic applications under real communication constraints. From a practical research perspective, this strict separation of concerns proved invaluable. Because communication support, workload placement, and resource isolation are handled by the lower layers, researchers were able to isolate 5G networking issues without modifying the ROS~2 application logic. Furthermore, during the initial setup of the FAST\_LIO pipeline~\citep{xu2021_fast_lio}, the team utilized container snapshots to safely trial different parameter configurations and ROS~2 distributions ensuring a stable and reproducible development process before the final execution.

\subsection{Use Case 2: Distributed GPU-Accelerated Semantic Mapping}
\label{subsec:use_case_2}

This second scenario addresses the construction of a 3D semantic map of an indoor environment, a representation that enriches geometric information with objects' instances and their categories~\cite{raul2017kbs}. In simplified terms, this process requires the concurrent execution of robot localization, sensor data pre-processing, object detection on RGB images, and the fusion of these detections and maintenance of the semantic map. These tasks require different software and hardware resources, which may be incompatible with each other. For example, they may depend on different library or CUDA versions, while some stages require GPU resources that may not be available on the robot itself. Therefore, this scenario is suitable to evaluate how CSAR supports the development of distributed robotic applications with heterogeneous requirements.

\subsubsection{Experimental Setup}
\begin{figure}
    \centering
    \includegraphics[width=1\linewidth]{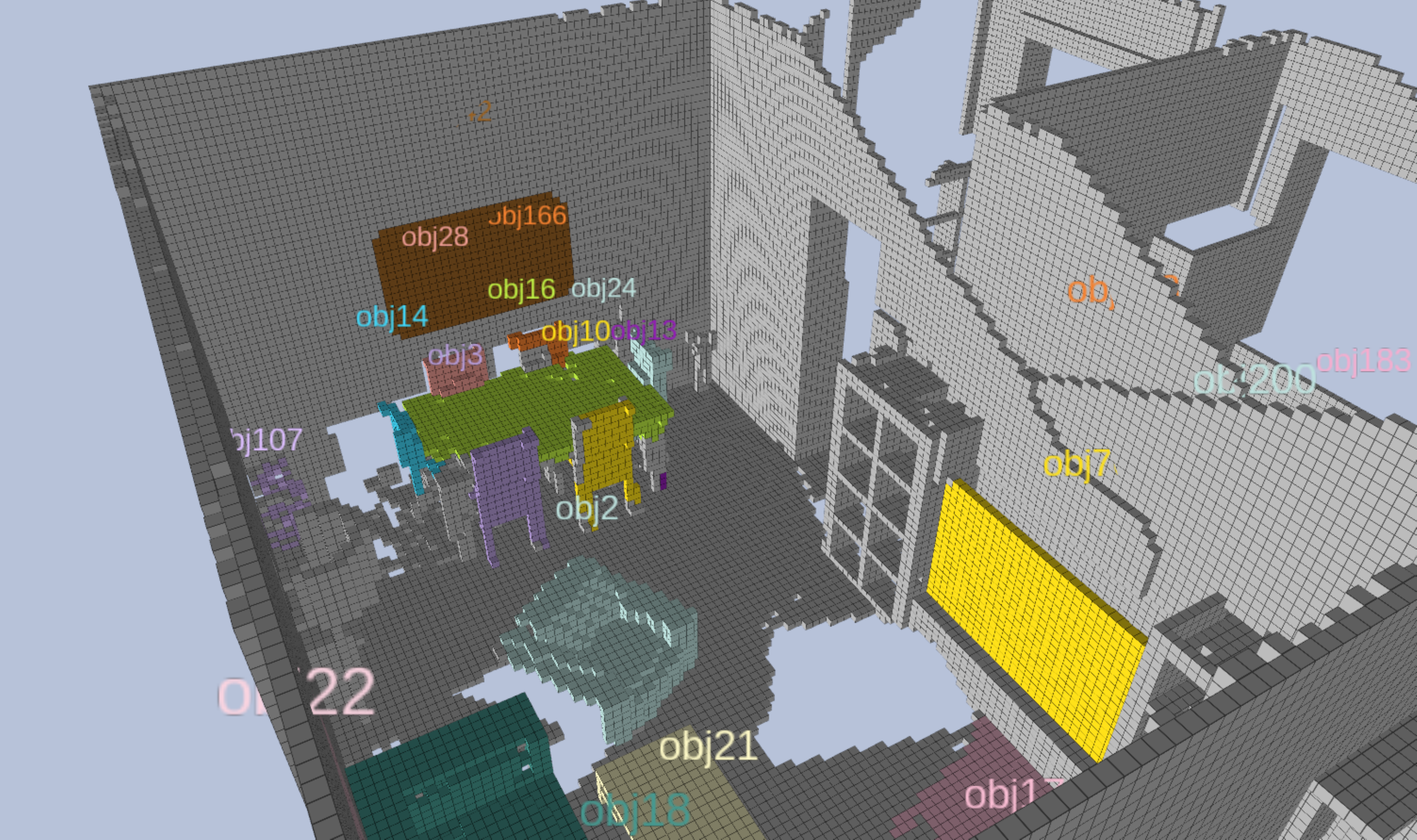}
    \caption{Voxelized 3D representation of the environment with object detections obtained from the Voxeland-based semantic mapping pipeline.}
    \label{fig:voxeland}
\end{figure}

\begin{figure*}
\centering
    \centering
    \includegraphics[width=1\linewidth]{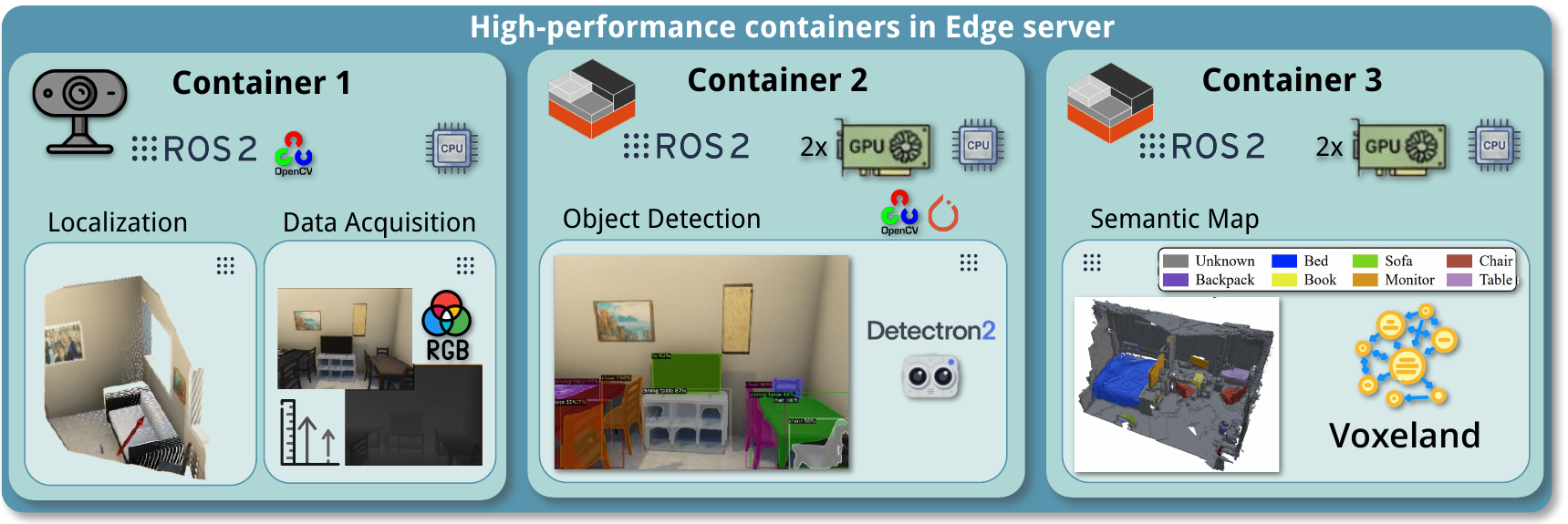}
    \caption{Distributed semantic mapping application decomposed into three CSAR system containers. Container~1, deployed on the robot, performs data acquisition and localization. Container~2, deployed on the edge server, executes the Detectron2-based object detection pipeline with GPU support. Container~3, also deployed on the edge server, runs Voxeland to build a probabilistic instance-aware semantic map of the environment. Communication among containers is performed through native ROS~2 topics and services.}
    \label{fig:CSAR_USE_CASE_2}
\end{figure*}

To implement this capability, the application was decomposed into three isolated but interconnected system containers executing ROS~2 nodes (see \FIG{\ref{fig:CSAR_USE_CASE_2}}):

\begin{itemize}

\item \textbf{Container 1 (Data Acquisition and Processing):}
This container is responsible for sensor data acquisition and pre-processing. In particular, it estimates the robot pose from the sensor observations and extracts geometric information from the environment, such as point clouds. This application only requires CPU resources and can therefore be executed locally on the robot.
\item \textbf{Container 2 (Object Detection):}
This container processes the RGB images received from Container~1 by using the Detectron2 object detection framework~\cite{wu2019_detectron2}. The network produces instance masks, object categories, and confidence values for the detected objects. This semantic information is then sent back to Container~1 for subsequent processing. Since this stage requires GPU resources, it is deployed on the \textit{Edge} server. 
\item \textbf{Container 3 (Semantic Map Management and Visualization):}
This container receives each new observation from Container~1, including both the robot pose and the geometric information already annotated with semantic information about the detected objects. It runs \textit{Voxeland} \cite{matez-bandera2024_voxeland}, which manages a voxelized 3D representation of the environment where the acquired knowledge is stored and progressively refined through the fusion of new observations (see \FIG{\ref{fig:voxeland}}). This task requires high CPU usage and may optionally benefit from GPU resources depending on the chosen perception and fusion configuration.
\end{itemize}

\subsubsection{Execution and Results}
This use case illustrates how CSAR supports the execution of distributed ROS~2 applications with different software and hardware requirements. In the implemented scenario, each container could be configured with the specific software versions required by its corresponding task, thus avoiding incompatibility problems such as conflicts between CUDA versions. In addition, CSAR managed the communication among the physical machines and the containers deployed on them, allowing the ROS~2 nodes to communicate in a natural way even when they were running on different hosts. Overall, the experiment validated the proposed approach for a semantic mapping task that originally presented difficulties related to software incompatibilities and inefficient GPU usage.

Beyond execution, these isolated containers served as persistent workspaces for the  team during the preparation of the experiment. Because each container operates as an independent Linux system, team members utilized their administrative privileges to actively write, compile, and debug their respective ROS~2 nodes directly. For instance, during the development of Container~2, researchers could iteratively test different versions of CUDA and the Detectron2 framework without risking system conflicts or disrupting the localization algorithms being simultaneously developed in Container~1. Furthermore, the ability to take immediate container snapshots allowed the team to safely rollback the environment during experimental trials, demonstrating that CSAR seamlessly accommodates both the setup and validation phases.

From an architectural perspective, this use case highlights the complementarity of the three CSAR layers. Layer~0 provides the shared computational infrastructure, including the GPU-equipped edge resources on which the most demanding stages can be executed. Layer~1 provides the isolated container environments and communication support required to interconnect heterogeneous ROS~2 components without exposing users to the complexity of the underlying infrastructure. Layer~2 hosts the application workloads themselves, allowing each stage of the semantic mapping pipeline to be configured with the specific software and hardware resources it requires. This decomposition is particularly valuable in multi-user settings, where different robotics team members may need to share accelerators and infrastructure while preserving independent software environments. Moreover, the monitoring mechanisms described in Section~\ref{subsec:monitoring} are especially relevant for this type of workload, where GPU utilization and container-level resource consumption must be observed to support stable shared operation.

\subsection{Additional Laboratory Uses}
\label{subsec:additional_uses}

Beyond the two representative use cases described above, CSAR has also been successfully used in other research scenarios within the laboratory. One example is a large-model-based semantic mapping pipeline~\citep{moncada-ramirez2024_large_models}, in which different stages of the mapping process combine visual perception, semantic description generation, large-model reasoning, and map exploitation. Another example is a multimodal human--robot interaction system~\citep{canete2024_multimodal_system}, where stereo audio processing, face detection and identification, and robot orientation towards its interlocutor are integrated into a distributed ROS~2 application. 

These additional cases are relevant because they broaden the spectrum of workloads supported by the framework. The former combines computationally demanding perception and reasoning components, whereas the latter requires the coordinated execution of sensing, control, and interaction modules under real-time constraints. 

Crucially, these advanced use cases were developed entirely within the CSAR infrastructure. The ability to provision full, root-privileged Linux containers allowed researchers to rapidly prototype and evaluate cutting-edge, experimental libraries, such as heavy LLM dependencies or novel audio processing tools, without the risk of breaking the host system or interfering with other users in the laboratory. In both cases, the corresponding works report that utilizing distributed Linux-container infrastructures not only improved the use of edge resources during execution,  but significantly simplified the preliminary integration of the robotic systems by preventing software dependency conflicts. These applications further illustrate the versatility of CSAR as an unified environment for both robotic experimentation and execution.

\section{Conclusions and Future Work}

This paper presented CSAR, a containerized architectural framework for robotics, designed to support the development, integration, deployment, and long-term operation of distributed robotic applications across heterogeneous computational resources. CSAR combines LXC/LXD-based system containerization, ROS~2-based communication, and a multi-layer edge infrastructure to address recurring challenges in modern robotic software, including dependency isolation, resource sharing, deployment portability, and the integration of compute-intensive workloads with mobile platforms. Through the description of the architectural framework, the real deployment in our laboratory, and the representative use cases, we illustrated that CSAR supports robotic workflows in which sensing, communication, and high-performance processing must coexist under practical infrastructure constraints. 
In particular, the results highlight the value of CSAR as a multi-user operating model, allowing robotics team members to share infrastructure and accelerators while preserving isolated and reproducible software environments. Beyond the deployment of final robotic tasks, CSAR fundamentally improves the preliminary experimentation, algorithm tuning, and integration phases for team members. Overall, CSAR provides a practical basis for organizing robotic computation across physical and virtual nodes while preserving experimental flexibility, reproducibility, and incremental system evolution.

In practical terms, the deployment in the MAPIR laboratory shows that the proposed framework enables safe prototyping of new robotic pipelines on shared infrastructure, controlled sharing of GPUs and sensing resources among multiple users, and reproducible long-term evolution of complex experiments without disrupting ongoing work.

As future work, we plan to extend CSAR in two directions that follow naturally from its current containerized, WAN-enabled design. 

First, we will study CSAR as a \textbf{policy-governed federation of ROS~2 domains}, where each local environment remains isolated by default and only selected topics, services, and actions are exposed across domains through DDS Router over the existing overlay network. We will evaluate scalability and robustness under representative WAN conditions, including latency, jitter, packet loss, intermittent connectivity, and mobility-driven link changes, as well as recovery after disruptions. 

Second, we will incorporate a \textbf{reproducible data plane} for robotics artifacts by combining object storage (e.g., MinIO/S3) for large unstructured data such as rosbags and video with a structured metadata layer (e.g., PostgreSQL/PostGIS) to index runs, trajectories, derived products, and model versions. The goal is to prevent experimental data from remaining scattered across isolated files and locations, where it becomes difficult to know later which run produced them, with which software version, and under which conditions. This will support traceable experiment storage, replay, and retrieval, including spatial queries, with ingestion and query performance characterized through repeatable benchmarks.

\section*{Acknowledgements}
This work has been supported by the projects MINDMAPS (PID2023-148191NB-I00) and Voxeland (PPRO-B1-2023-017, JA.B1-09), funded by the Ministry of Science, Innovation, and Universities of Spain and by the University of Málaga, respectively, as well as by the Andalusian Plan for Research, Development, and Innovation (PPRO-TEP960-G-2023). Funding for open access charge: Universidad de Málaga / CBUA. We also thank all MAPIR members who contributed to the development of CSAR and facilitated the preparation of this paper.
%

%
%

During the preparation of this work the author(s) used Perplexity in order to improve English writing style.

\bibliographystyle{elsarticle-num}
\bibliography{csar}

\end{document}